\documentclass[a4paper,fleqn]{cas-sc}

\usepackage[authoryear]{natbib}
\usepackage{float}

\usepackage{subfigure}
\usepackage{amsthm}
\usepackage[section]{placeins} 
\usepackage{array}
\usepackage{amsmath}  
\usepackage{algorithm}
\usepackage{algpseudocode}
\usepackage[dvipsnames]{xcolor}
\usepackage[most]{tcolorbox}
\usepackage{tabularx}

\usepackage{tcolorbox} % For tcolorbox environment
\usepackage{listings}  % For lstlisting environment

\def\tsc#1{\csdef{#1}{\textsc{\lowercase{#1}}\xspace}}
\tsc{WGM}
\tsc{QE}
\tsc{EP}
\tsc{PMS}
\tsc{BEC}
\tsc{DE}
\hypersetup{
  colorlinks,
  citecolor=blue,
  linkcolor=blue,
  urlcolor=blue}
\begin{document}
\let\WriteBookmarks\relax
\def\floatpagepagefraction{1}
\def\textpagefraction{.001}

\definecolor{lblue}{rgb}{0.12, 0.57, 1.0}

\shorttitle{LD-Scene: LLM-Guided Diffusion for Controllable Generation of Adversarial Safety-Critical Driving Scenarios}

\shortauthors{M. PENG}

\title [mode = title]{LD-Scene: LLM-Guided Diffusion for Controllable Generation of Adversarial Safety-Critical Driving Scenarios}

\author[1]{Mingxing Peng}
\ead{mpeng060@connect.hkust-gz.edu.cn}

\author[2]{Yuting Xie}
\ead{xieyt8@mail2.sysu.edu.cn}

\author[1]{Xusen Guo}
\ead{xguo796@connect.hkust-gz.edu.cn}

\author[1]{Ruoyu Yao}
\ead{ryao092@connect.hkust-gz.edu.cn}

\author[1, 3]{Hai Yang}
\ead{cehyang@ust.hk}

\author[1, 3]{Jun Ma}
\ead{jun.ma@ust.hk}
\cormark[1]

\affiliation[1]{organization={The Hong Kong University of Science and Technology (Guangzhou)},
    % postcode={511453}, 
    city={Guangzhou 511453},
    country={China}}

\affiliation[2]{organization={School of Computer Science and Engineering, Sun Yat-sen University},
    % postcode={510275}, 
    city={Guangzhou 510275},
    country={China}
    }
\affiliation[3]{organization={The Hong Kong University of Science and Technology},
    city={Hong Kong SAR},
    country={China}}

\cortext[cor1]{Corresponding author}

\begin{abstract}
Ensuring the safety and robustness of autonomous driving systems necessitates a comprehensive evaluation in safety-critical scenarios. 
However, these safety-critical scenarios are rare and difficult to collect from real-world driving data, posing significant challenges to effectively assessing the performance of autonomous vehicles. Typical existing methods often suffer from limited controllability and lack user-friendliness, as extensive expert knowledge is essentially required. 
To address these challenges, we propose LD-Scene, a novel framework that integrates Large Language Models (LLMs) with Latent Diffusion Models (LDMs) for user-controllable adversarial scenario generation through natural language.
Our approach comprises an LDM that captures realistic driving trajectory distributions and an LLM-based guidance module that translates user queries into adversarial guidance functions, facilitating the generation of scenarios aligned with user queries. 
The guidance module integrates an LLM-based Chain-of-Thought (CoT) code generator and an LLM-based code debugger, enhancing the controllability and robustness in generating guidance functions. 
Extensive experiments conducted on the nuScenes dataset demonstrate that LD-Scene achieves state-of-the-art performance in generating realistic and effective adversarial scenarios. Furthermore, our framework provides fine-grained control over adversarial behaviors, thereby facilitating more effective testing tailored to specific driving scenarios.
\end{abstract}

% Keywords
% Each keyword is separated by \sep

\maketitle

\section{Introduction}\label{sec:1}

With the continuous advancement of autonomous vehicle (AV) technologies, reliable testing is essential to verify the safety and robustness of self-driving systems \citep{10591480, 10716596, kang2019test}. However, real-world testing is not only expensive but also poses significant challenges in collecting safety-critical driving scenarios, which are crucial for assessing the performance of AVs. Specifically, safety-critical scenarios are characterized by high-risk interactions where AVs are exposed to unexpected maneuvers by other road users, which potentially lead to collisions \citep{ding2023survey, 10844066, 10779443}. As a more practical and efficient solution, the generation of safety-critical scenarios in simulation has become a widely adopted approach for driving performance evaluation.

Existing approaches often employ simulators such as CARLA \citep{dosovitskiy2017carla} and SUMO \citep{lopez2018microscopic} to manually design safety-critical driving scenarios. However, these methods typically demand substantial domain expertise and often produce scenarios that lack realism. Recent studies have sought to address these limitations by leveraging large-scale driving datasets to learn realistic traffic models, subsequently generating safety-critical scenarios at test time through optimization-based techniques \citep{strive, advsim, advtraj}. While these approaches improve realism to some extent, they remain limited in terms of generation efficiency and fine-grained behavioral control. More recently, diffusion models have demonstrated strong capabilities in modeling complex trajectory distributions through iterative denoising steps \citep{mao2023leapfrog, peng2024diffusion}. Moreover, diffusion models have also been widely adopted in controllable generation tasks, such as text generation \citep{li2022diffusion} and image synthesis \citep{zheng2023layoutdiffusion}, attracting considerable interest for their potential to generate realistic and controllable driving trajectories. Existing diffusion-based approaches employ various guidance mechanisms, such as reinforcement learning (RL)-based classifiers \citep{xie2024advdiffuser} and predefined objective functions \citep{xu2023diffscene, chang2024safe}, to generate adversarial safety-critical scenarios. However, these approaches require retraining classifiers or redesigning objective functions with expert knowledge for different adversarial strategies, and these limit their flexibility and user-friendliness. With the above discussions, the key challenges in generating safety-critical scenarios lie in ensuring realism, achieving controllability over adversarial behaviors, and providing a user-friendly interface for scenario generation.

Recent advancements in generative AI provide new solutions to deal with these challenges. In particular, Large Language Models (LLMs) have demonstrated remarkable capabilities in natural language understanding \citep{achiam2023gpt, bubeck2023sparks} and code generation \citep{guo2024deepseek}, making them particularly suitable for developing interfaces that allow users to customize scenarios in natural language. 
For example, CTG++ \citep{zhong2023language} leverages language-based guidance to enable controllable generation of multi-agent traffic simulations. However, it lacks closed-loop evaluation of an ego planner and does not support reactive, adversarial safety-critical scenarios where an attacking agent challenges the planner's decisions. In addition, the inherent instability of LLMs leads to frequent failures in generating valid guidance code, undermining the reliability of CTG++. 
Meanwhile, Latent Diffusion Models (LDMs) \citep{rombach2022high} have gained attention for their ability to learn compact representations of complex driving scenarios, which facilitate more realistic and computationally efficient trajectory modeling. These advancements enhance both the expressiveness and efficiency of scenario generation. Inspired by these developments, integrating multi-agent LLM with LDMs presents a promising approach for generating realistic and controllable adversarial safety-critical driving scenarios while enabling intuitive and accessible user interaction.

In this paper, we propose LD-Scene, a novel approach that integrates LLM-enhanced guidance with LDMs to enable user-controllable generation of adversarial safety-critical driving scenarios through natural language. As illustrated in Fig.~\ref{fig:overall_framework}, our LD-Scene framework consists of two main components: an LDM, which learns realistic driving trajectories, and an LLM-based guidance generation module, which translates user queries into a guidance loss function that perturbs the denoising process to generate adversarial driving scenarios. Specifically, both past information and future information are encoded into latent representations by a graph neural network (GNN)-based encoder, which captures vehicle interactions. The past latent serves as conditions for the denoising network to reconstruct a realistic future latent, ensuring the model learns plausible driving behaviors. The guidance generation module consists of a code generator and a code debugger. The Chain-of-Thought (CoT) reasoning process within the code generator enhances the controllability of adversarial scenario generation, allowing not only the specification of adversarial behaviors but also the adjustment of adversarial intensity. Meanwhile, the code debugger mitigates the instability inherent in LLM-based generation, thereby improving the robustness and effectiveness of the proposed approach. 
We conduct extensive closed-loop simulation experiments with an ego vehicle controlled by a rule-based planner to evaluate the effectiveness and realism of the generated scenarios. In addition, ablation studies validate the contribution of the debugger module to the stability of LD-Scene.

In summary, the main contributions of this paper include: 
\begin{itemize}
    \item We propose LD-Scene, a novel framework that seamlessly integrates the multi-agent LLM with an LDM to facilitate the generation of adversarial, safety-critical driving scenarios that are effortlessly controllable through natural language.

    \item We introduce an LLM-based guidance generation module that features a CoT reasoning code generator and a code debugger, and this enhances the controllability, robustness, and stability of adversarial scenario generation for autonomous driving.

    \item Extensive experiments on the nuScenes dataset are conducted, and the results demonstrate that LD-Scene outperforms baseline models in terms of adversariality, realism, and stability, while also providing improved controllability over both the adversarial level and specific adversarial behaviors.
        
\end{itemize}

\section{Related Work}
\label{sec:related work}
This section introduces related works in three areas: safety-critical scenario generation, diffusion models for trajectory generation, and LLM-based traffic simulation.

\subsection{Safety-Critical Driving Scenario Generation} 
Simulating safety-critical scenarios is essential for comprehensive risk assessment of autonomous driving systems, especially since current AVs have been shown to perform well in typical driving conditions but remain under-tested in rare and hazardous cases~\citep{ding2023survey}. Manual design of safety-critical scenarios, relying on expert knowledge and altering factors like actor locations and velocities, faces scalability issues and may yield implausible situations~\citep{waymo, metadrive, closed-loop}.
Recent studies delve into some specific parameterization spaces within original scenarios to pinpoint adversarial parameters using optimization-based methods~\citep{advsim, advtraj,advsce,RN246,RN247, RN249,RN250}. Most works like AdvSim~\citep{advsim} directly perturb the standard trajectory space to produce adversarial trajectories, while adhering to constraints to maintain physical feasibility. Alternatively, some other works~\citep{RN246, mixsim, strive} perform parameter optimization within a condensed latent space. Strive~\citep{strive} further incorporates GNNs~\citep{scarselli2008graph} to generate traffic prior, while Mixsim~\citep{mixsim} utilizes routes as priors both to enhance the plausibility of vehicle interactions. Nevertheless, these testing-time optimization methods encounter practical constraints that necessitate the iterative re-planning of a surrogate planner within the search loop, resulting in significant efficiency issues.

Other studies adopt adversarial policy models to address the sequential control of BVs in step-wise interactions. Building upon Proximal Policy Optimization (PPO) \citep{ppo}, NADE~\citep{ade} trains background vehicles to execute adversarial maneuvers through simulation with a surrogate target model, offering flexibility and efficiency in scenario generation. However, this approach introduces complexity due to the curse of dimensionality and rarity. To alleviate this issue, D2RL~\citep{drl} proposes a strategy to enhance information density by training neural networks with safety-critical data. They also selectively choose a critical state attacker to further reduce variance, sacrificing multi-vehicle trajectory rationality. The commendable efforts notwithstanding, adversarial policy training still necessitates a substantial number of interactions with the environment due to the inherent model complexity. 

\begin{figure*}[!h]
\centerline{\includegraphics[width=\linewidth, trim= 45 150 250 0, clip]{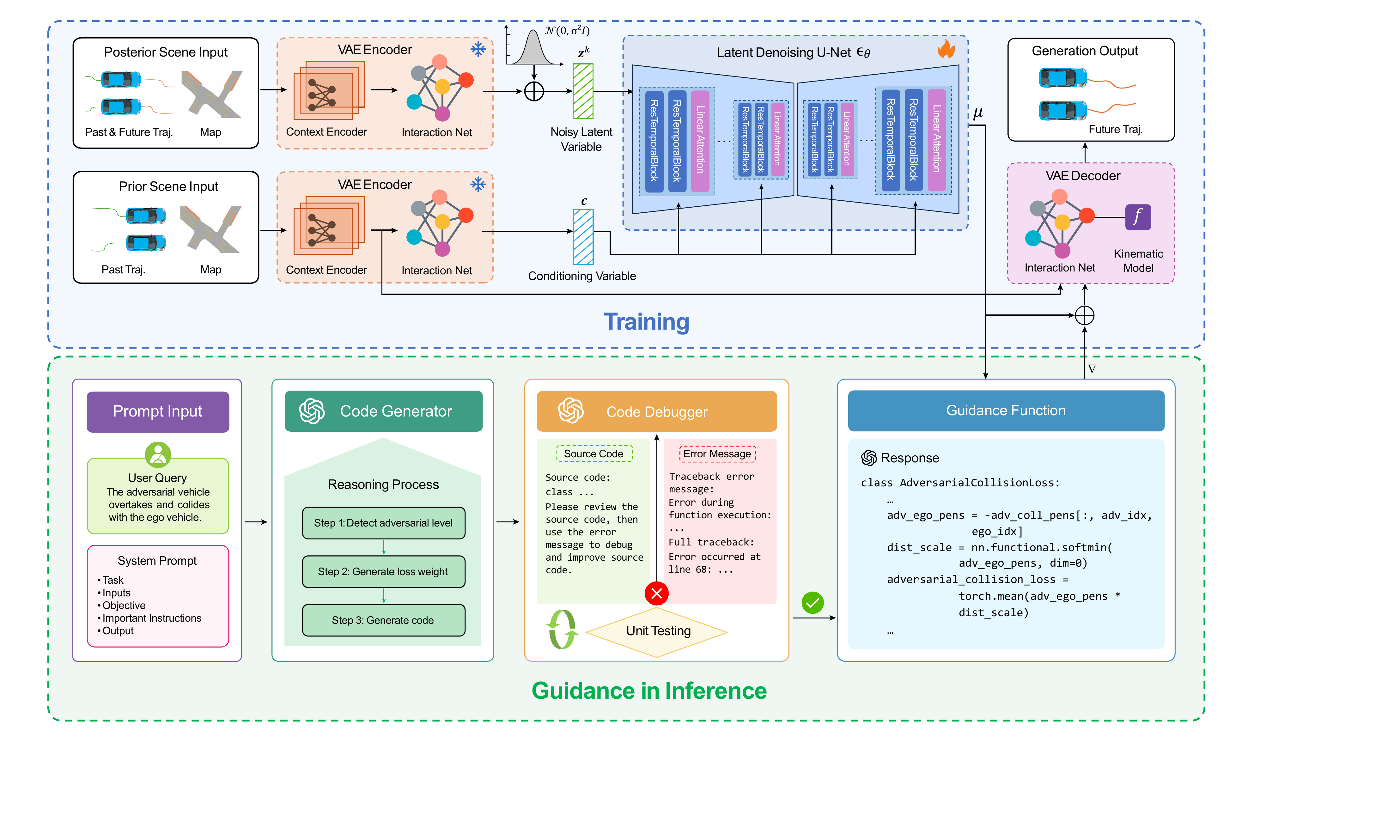}}
\caption{Overall framework of LD-Scene. During the training stage, an LDM learns the distribution of realistic driving trajectories conditioned on the latent representation of historical scene input. During the inference stage, given a user query, an LLM-based code generator produces an adversarial loss function. This loss function is then validated by an LLM-based debugger through a closed-loop unit testing process and subsequently used to guide the diffusion model in generating safety-critical driving scenarios.}
\label{fig:overall_framework}
\end{figure*}

\subsection{Diffusion Models for Controllable Trajectory Generation}
Diffusion models \citep{ho2020denoising, rombach2022high, dhariwal2021diffusion} have emerged as a powerful class of generative models, attracting significant attention for their ability to generate realistic and controllable driving trajectories. Their capacity to effectively capture the complexity of real-world traffic patterns makes them well-suited for traffic simulation tasks \citep{peng2024diffusion}. Furthermore, guidance-based diffusion models improve controllability at test time while maintaining realism \citep{jiang2024survey}. For example, Diffuser \citep{janner2022planning} and AdvDiffuser \citep{xie2024advdiffuser} leverage RL to train a reward function, which acts as a classifier to guide trajectory sampling. CTG \citep{zhong2023guided} employs Signal Temporal Logic (STL) formulas as guidance for diffusion models to ensure compliance with specific rules, such as collision avoidance or goal-reaching. CTG++ \citep{zhong2023language} further incorporates language-based guidance, enabling more flexible rule enforcement. MotionDiffuser \citep{jiang2023motiondiffuser} proposes several differentiable cost functions as guidance, enabling the enforcement of both rules and physical constraints in the generated trajectories. In contrast, DiffScene \citep{xu2023diffscene} and Safe-Sim \citep{chang2024safe} focus on generating safety-critical traffic simulations by introducing safety-based guidance objectives. Furthermore, Safe-Sim \citep{chang2024safe} demonstrates the ability to generate controllable behaviors, such as specific collision types. 

Despite the successes achieved in the controllable generation of realistic driving trajectories, these prior works still face certain limitations in terms of user convenience and controllable generation efficiency. For example, RL-based guidance, such as AdvDiffuser \citep{xie2024advdiffuser}, requires the training of different classifiers for various adversarial strategies. On the other hand, the controllability achieved through pre-designed objective-based guidance, such as Safe-Sim \citep{chang2024safe}, often demands significant domain expertise, making it rather difficult for users without specialized knowledge in the field.

\subsection{Large Language Models for Traffic Simulation}
Recently, LLMs have represented a significant advancement in the field of autonomous driving \citep{yang2023llm4drive, peng2024lc, wen2023dilu}, showcasing their potential in traffic simulation for generating realistic and controllable scenarios. This progress is largely driven by their capabilities in extensive knowledge storage, logical reasoning, and code generation \citep{bubeck2023sparks, guo2024deepseek}. CTG++ \citep{zhong2023language} and \cite{guo2024automating} demonstrate LLM's powerful code generation capabilities that can generate corresponding functions according to user requirements. Meanwhile, Scenediffuser \citep{jiang2024scenediffuser} employs LLMs through few-shot prompting to convert scene constraints into structured Protobuf (Proto) representations, facilitating controllable scenario generation. LCTGen \citep{tan2023language} employs LLMs as interpreters to transform textual queries into structured representations, which are subsequently combined with a transformer-based decoder to generate realistic traffic scenes. Building upon LCTGen, an enhanced method is introduced in \cite{xia2024language}, where the structured representations produced by LLMs incorporate interaction modeling, ultimately enhancing the generation of interactive traffic trajectories. On the other hand, ChatScene \citep{zhang2024chatscene} utilizes an LLM-based agent to describe traffic scenes in natural language, subsequently converting these descriptions into executable Scenic code for scenario simulation within CARLA. Although these works have demonstrated a certain degree of controllability in traffic simulation, few studies have specifically focused on adversarial safety-critical scenario generation. In this context, it leaves an open and interesting problem to integrate LDMs with LLMs to achieve both realism and user-friendly controllability in the generation of safety-critical traffic scenarios.

\section{Methodology}
\label{sec:methodology}
In this section, we introduce LD-Scene, a framework that integrates LDMs with LLM-enhanced guidance to generate plausible safety-critical driving scenarios with user-friendly controllability. The overall framework of our LD-Scene is depicted in Fig.~\ref{fig:overall_framework}. The following subsections provide a formal problem formulation, describe the LDM for scenario generation, detail the process of generating LLM-enhanced guidance, and explain how this guidance is utilized to generate safety-critical driving scenarios.

\subsection{Problem Formulation}
Our work focuses on generating safety-critical driving scenarios for a given autonomous driving planner to facilitate a more effective and efficient evaluation of autonomous driving systems. Each scenario consists of $ N$ agents, including an ego vehicle controlled by the planner $\pi$, while the future trajectories of the remaining $N-1$ vehicles are generated by our model. Among the $N-1$ vehicles, one is designated as the adversarial vehicle. The objective of our model is to encourage the adversarial vehicle to induce a collision with the ego vehicle while ensuring that the trajectories of the remaining non-adversarial vehicles remain realistic. Furthermore, our model incorporates LLMs to enable scenario customization based on user queries in natural language. This allows users to conveniently specify the adversarial vehicle, as well as control the collision type and adversarial collision severity.

Similar to prior work \citep{strive, xie2024advdiffuser}, a driving scenario $S$ consists of $N$ vehicle states and a map $\textit{\textbf{m}}$ that includes semantic layers for drivable areas and lanes. At any timestep $t$, the states of the $N$ agents are represented as $s_{t} = [s^{0}_{t}, s^{1}_{t}, ..., s^{N-1}_{t}]$, where each agent state $s^{i}_{t} = (x^{i}_{t}, y^{i}_{t}, \theta^{i}_{t}, v^{i}_{t})$ includes the 2D position, heading, and speed. The corresponding actions of the $N$ agents are represents as $a_{t} = [a^{0}_{t}, a^{1}_{t}, ..., a^{N-1}_{t}]$, where each agent action $a^{i}_{t} = (\dot{v}_i^t, \dot{\theta}_i^t)$ representing the acceleration and yaw rate. The past trajectories, denoted as $\textit{\textbf{x}}$, represent the historical states of all agents over the past $T_{hist}$ timesteps and are expressed as $\textit{\textbf{x}} = \{s_{t-T_{hist}}, s_{t-T_{hist}+1}, ..., s_{t}\}$. The planner $\pi$ determines the future trajectory of the ego vehicle over a time horizon from $t$ to $t + T$. The planned future trajectory is denoted as $s^{0}_{t:t+T} = \pi (\textit{\textbf{m}}, \textit{\textbf{x}})$, where $\pi$ processes historical state sequences and map feature to predict a sequence of future states.

Our proposed LDM $g$, parameterized by $\theta$, is designed to generate multi-agent future trajectories in a driving scenario. The model consists of a pretrained encoder $\mathcal{E}$ and decoder $\mathcal{D}$. Given historical trajectory data and corresponding map information, the encoder $\mathcal{E}$ maps these inputs into a compact latent representation. A denoising diffusion process is then applied within this latent space, gradually refining the representation. Finally, the decoder $\mathcal{D}$ transforms the denoised latent representation into future trajectories of non-ego vehicles, denoted as $\boldsymbol{\tau} = \{s^{i}_{t:t+T}\}^{N-1}_{i=1}$. In the training stage, the model is trained on real-world driving data to learn realistic traffic behaviors. In the inference stage, we incorporate LLM-enhanced guidance to steer the denoising sampling process toward the generation of safety-critical scenarios, thereby ensuring both realism and controllability.

\begin{figure*}[!h]
\centerline{\includegraphics[width=0.9\linewidth, trim= 0 20 0 0, clip]{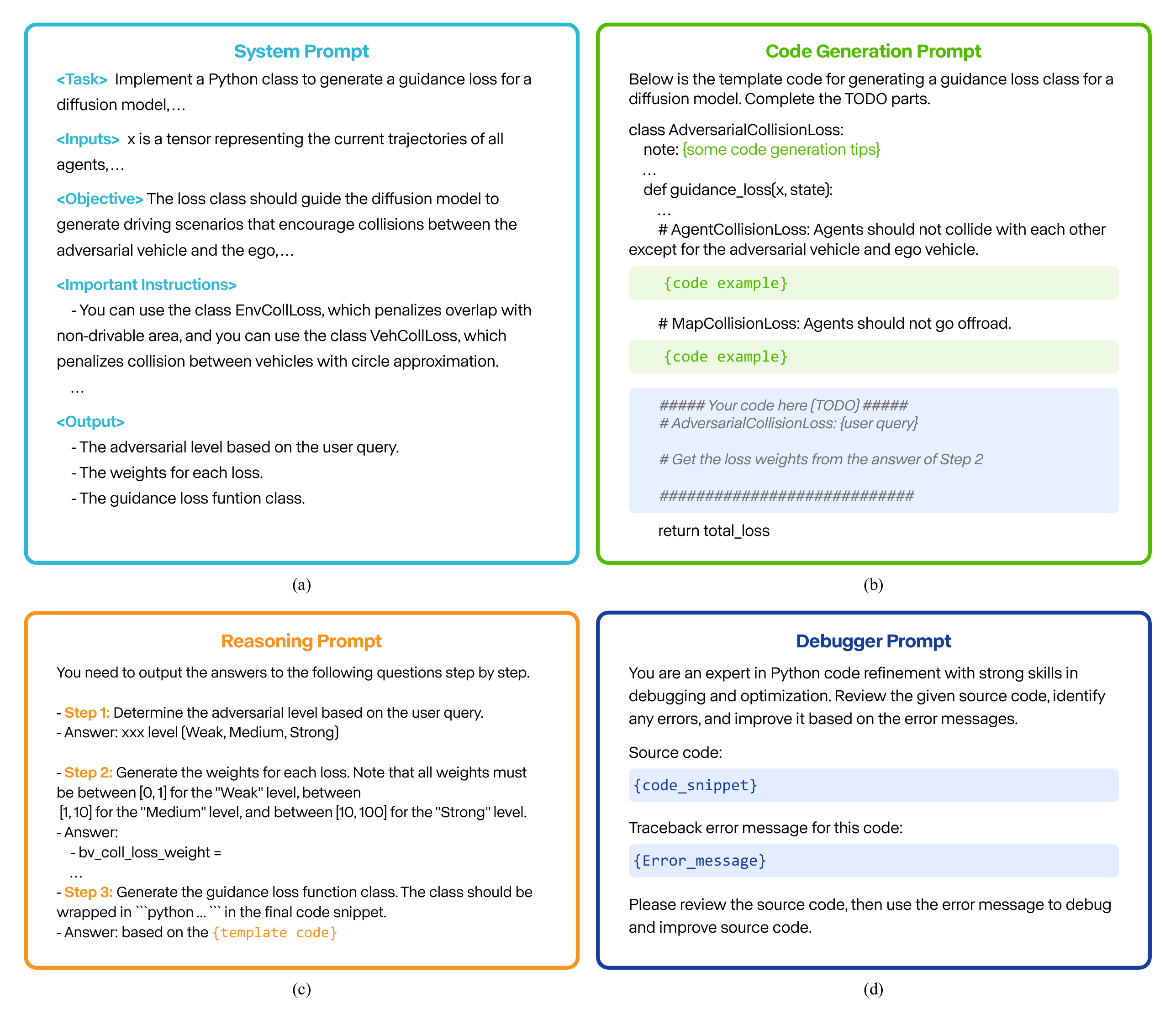}}
\caption{Prompts in the guidance generation module. (a) System Prompt: it specifies the task, inputs, objectives, and important instructions for generating the guidance loss function. (b) Code Generation Prompt: it provides a structured template for generating a guidance loss, incorporating predefined loss functions and some code generation tips. (c) Reasoning Prompt: it outlines a step-by-step reasoning process to determine the adversarial level, assign appropriate loss weights, and generate the guidance loss function. (d) Debugger Prompt: it instructs the model to analyze generated source code and error messages and iteratively refine the implementation to improve correctness and reliability.}
\label{fig:prompt}
\end{figure*}

\subsection{Latent Diffusion Models for Scenario Generation}
Diffusion models \citep{ho2020denoising} comprise two Markov chains: a forward (diffusion) process that progressively adds Gaussian noise to the data, leading to pure noise over multiple steps, and a reverse (denoising) process, where a learnable neural network iteratively removes noise to generate samples. In this work, we train a diffusion model to synthesize realistic and controllable adversarial safety-critical driving scenarios through an iterative denoising process.

Unlike conventional diffusion-based trajectory generation methods that operate directly in the trajectory space \citep{zhong2023guided, zhong2023language, chang2024safe}, our approach applies the diffusion model within a latent space. This design choice reduces computational complexity while enhancing the expressiveness of the generative process \citep{rombach2022high, strive, xie2024advdiffuser}. Specifically, we integrate the diffusion model with a pretrained graph-based variational autoencoder (VAE) model following Strive~\citep{strive}. The graph-based VAE consists of an encoder that learns a latent representation of agent interactions and a decoder that autoregressively generates future trajectories using a kinematic bicycle model, ensuring both realism and plausibility in traffic scenario generation.

\textbf{Architecture.} 
As depicted in the upper section of Fig.~\ref{fig:overall_framework}, our LDM consists of three key components: two encoders with frozen parameters, a learnable denoising network, and an autoregressive trajectory decoder with frozen parameters. The encoders and decoder are directly adopted from a pretrained VAE model, which all these modules employ GNNs. Specifically, the prior scene input encoder $\mathcal{E}_\theta$ is identical to the prior network in Strive \citep{strive}, modeling agent interactions via a fully connected scene graph. Each node encodes contextual features derived from an agent's past trajectory and local rasterized map. Through message passing, the encoder generates latent representations for all agents, which serve as conditioning inputs for our diffusion model, formulated as $\textit{\textbf{c}}=\mathcal{E}_\theta(\textit{\textbf{x}}, \textit{\textbf{m}})$. Similarly, the posterior scene input encoder $\mathcal{E}_\theta$ corresponds to the posterior network in Strive \citep{strive}, which operates jointly on past and future information. The resulting latent vectors serve as the latent input $\mathbf{z}$ for our diffusion model, expressed as $\mathbf{z} = \mathcal{E}_\theta(\boldsymbol{\tau}, \textit{\textbf{x}}, \textit{\textbf{m}})$. The diffusion process begins with a clean future latent $\mathbf{z}^0 \sim q(\mathbf{z}^0)$ sampled from the data distribution. The forward process produces a sequence of progressively noisier latent $(\mathbf{z}^0, \mathbf{z}^1, ..., \mathbf{z}^K)$, where each step $k$ follows a Gaussian noise injection process \citep{ho2020denoising}:
\begin{equation}
\begin{aligned}
q(\mathbf{z}^{1:K} \mid \mathbf{z}^0) &:= \prod_{k=1}^{K} q(\mathbf{z}^k \mid \mathbf{z}^{k-1}) \\
q(\mathbf{z}^k \mid \mathbf{z}^{k-1}) &:= \mathcal{N} \left(\mathbf{z}^k ; \sqrt{1 - \beta_k} \mathbf{z}^{k-1}, \beta_k \mathbf{I} \right)
\end{aligned}
\end{equation}
where $\beta_k$ is the predefined variance schedule that controls the noise level at each step. For sufficiently large $K$, the final latent variable $\mathbf{z}^K$ approaches an isotropic Gaussian distribution, i.e., $\mathcal{N}(\mathbf{0}, \mathbf{I})$. 

To accelerate the inference process, we employ the Denoising Diffusion Implicit Models (DDIM) sampling strategy \citep{song2020denoising}, which facilitates a non-Markovian formulation of the reverse diffusion process. Unlike conventional sampling schemes, DDIM enables efficient generation by skipping intermediate denoising steps, thus significantly reducing the number of sampling iterations without necessitating model retraining. The reverse process at step $k$ is defined as:
\begin{equation}
\mathbf{z}^{k-1} = \sqrt{\alpha_{k-1}} \cdot \tilde{\mathbf{z}}^0 + \sqrt{1 - \alpha_{k-1}} \cdot \epsilon_\theta(\mathbf{z}^k, k, \mathbf{c})
\label{denoising_process}
\end{equation}
where $\epsilon_\theta(\mathbf{z}^k, k, \textit{\textbf{c}})$ denotes the denoising network that predicts the noise conditioned on the step $k$ and conditioning latent $\textit{\textbf{c}}$. The coefficient $\alpha_k = \prod_{i=1}^{k} (1 - \beta_i)$ represents the cumulative product of the noise schedule. The estimate of the clean latent representation $\tilde{\mathbf{z}}^0$ at step $k$ is given by:
\begin{equation}
\tilde{\mathbf{z}}^0 = \left( \frac{\mathbf{z}^k - \sqrt{1 - \alpha_k} \cdot \epsilon_\theta(\mathbf{z}^k, k, \mathbf{c})}{\sqrt{\alpha_k}} \right)
\end{equation}
By iteratively applying this reverse denoising operation starting from the noisy latent $\mathbf{z}^K$, we obtain the final denoised sample $\hat{\mathbf{z}}^0$.

As illustrated in Fig.~\ref{fig:overall_framework}, the denoising network adopts a U-Net architecture similar to \cite{janner2022planning}, composed of one-dimensional temporal convolutional blocks with stacked residual blocks and linear attention. The conditioning input $\textit{\textbf{c}}$ is incorporated into the intermediate input latent within each convolutional block.

Finally, the trajectory decoder $\mathcal{D}$, following the architecture in Strive \citep{strive}, also employs a GNN-based interaction net to ensure realistic vehicle interactions. Specifically, the decoder $\mathcal{D}_\theta(\hat{\mathbf{z}}^0, \textit{\textbf{x}}, \textit{\textbf{m}})$ operates on the scene graph with both the denoised latent and past information embedding. The decoding process is autoregressive, where the model predicts all agents' actions $a_t$ at timestep $t$, which are then propagated through a kinematic bicycle model $f$ to compute the next state $s_{t+1}$. The updated state is incorporated into the embedding before proceeding to the next time step. This autoregressive formulation ensures the physical plausibility of the generated trajectories while maintaining realistic multi-agent interactions.

\textbf{Training.}  
Since the VAE is pretrained, we focus on training the LDM. The objective is to train a denoising network that minimizes the variational bound on the negative log-likelihood, which can be simplified to a mean squared error loss between the predicted noise and the actual noise:  
\begin{equation}
\mathcal{L} = \mathbb{E}_{\mathbf{z}^k, \epsilon \sim \mathcal{N}(0, I), k, \textit{\textbf{c}}} \left[ \|\epsilon - \epsilon_{\theta}(\mathbf{z}^k, k, \textit{\textbf{c}})\|^2 \right]
\end{equation}
where $\epsilon \sim \mathcal{N}(0, I)$ is the Gaussian noise, and $\mathbf{z}^k$ is the noisy latent input obtained from the encoder $\mathcal{E}$ with the forward diffusion process. $\epsilon_{\theta}(\mathbf{z}^k, k, \textit{\textbf{c}})$ is the noise prediction model, which is conditioned on the past context latent $\textit{\textbf{c}}$.  

\subsection{Guidance Generation with LLMs}
In this work, we propose a novel approach that leverages LLMs for the controllable generation of adversarial driving scenarios. Our framework takes a user query as input and utilizes LLMs to generate the code for a loss function, which subsequently serves as guidance for the sampling process, ensuring that the generated scenarios adhere to user-defined adversarial objectives. The key components of our guidance generation framework are illustrated in Fig.~\ref{fig:overall_framework} and consist of the inputs, an LLM-based code generator, and an LLM-based code debugger.

\subsubsection{User Query and System Prompt}
The natural language user query specifies an adversarial driving event. For instance, the query may describe a scenario in which an adversarial vehicle overtakes and collides with the ego vehicle. The structured system prompt defines the task, introduces input parameters, specifies objectives, provides essential code generation instructions, and outlines the expected outputs. Fig.~\ref{fig:prompt}(a) presents a detailed illustration of the system prompt.

\subsubsection{Code Generator}
CoT reasoning has been shown to be highly effective in problem decomposition and in carrying out more intricate reasoning tasks \citep{wei2022chain, kojima2022large}. Therefore, our code generator adopts CoT prompting to enhance effectiveness and provide greater controllability in adversarial scenario generation. The code generator prompts in our framework consist of two primary modules: the reasoning prompt and the code generation prompt. As shown in Fig.~\ref{fig:prompt}(b), the code generation prompt provides a template that ensures the correct implementation of the guidance loss function, while Fig.~\ref{fig:prompt}(c) presents the reasoning prompt, which instructs the model to generate the loss function step-by-step.

The Reasoning Module is divided into three steps. First, it interprets the user query to determine the adversarial level (Weak, Medium, or Strong). The LLM follows some instructions to classify the level: if the query contains terms with ambiguous intensity, it is categorized as Medium. Queries that include strong descriptors (e.g., aggressive, forceful, high-speed) are classified as Strong, whereas those with mild descriptors (e.g., gentle, cautious, slight) are categorized as Weak. Second, the module determines the appropriate loss weights based on predefined ranges, which are essential for controlling the adversarial level of the generated driving scenario. Third, the system generates the corresponding code by completing a predefined template, incorporating the determined adversarial level and loss weights to ensure effectiveness in adversarial scenario generation.

Code generation prompt. Unlike the few-shot learning approach used in CTG++ \citep{zhong2023language}, our method employs a zero-shot learning strategy for code generation. The Code Generation Prompt provides a structured template that directly incorporates examples of some loss functions, including the agent collision loss function and the map collision loss function, as shown in Fig.~\ref{fig:prompt}(b). Based on these examples, the LLM learns how to utilize the input trajectory and apply helper classes such as EnvCollLoss and VehCollLoss. Finally, the code generator generates the adversarial collision loss based on the user query. Depending on different user queries, such as cut-in, overtaking, or emergency braking, the corresponding adversarial collision loss function is generated, ensuring controllable adversarial behavior.

\subsubsection{Code Debugger}
As noted in CTG++ \citep{zhong2023language}, code generation can sometimes produce incorrect implementations. To address this limitation, we introduce a code debugger module that reviews and refines the generated code. The debugger operates by running a closed-loop unit test to evaluate the generated guidance loss function. If an error is detected, the corresponding source code and error message are provided to the code debugger, which iteratively refines the implementation until a predefined maximum number of iterations is reached.

Fig.~\ref{fig:prompt}(d) illustrates the debugging prompt, which instructs the model to analyze the generated source code, diagnose errors based on the traceback message, and improve the implementation. This iterative refinement process enhances code reliability, ensuring seamless integration of the guidance loss function into the diffusion model while minimizing the need for manual debugging. By incorporating a code debugger, our framework significantly improves the robustness and correctness of the generated loss functions, thereby enhancing the effectiveness of adversarial scenario generation.

\subsection{Generation of Safety-Critical Scenarios}
To enable the controllable generation of safety-critical driving scenarios, we introduce an objective function $\mathcal{J}(\boldsymbol{\tau})$, which guides the denoising process. Adversarial driving scenarios are generated by perturbing the predicted mean at each denoising step. Since the denoising process operates in the latent space, each guided iteration step first decodes the latent vector $\mathbf{z}$ into the corresponding trajectory $\boldsymbol{\tau}$ using a decoder $\mathcal{D}$, formulated as $\boldsymbol{\tau} = \mathcal{D}_\theta (\mathbf{z}, \textit{\textbf{x}}, \textit{\textbf{m}})$. At each reverse diffusion step $k$, we modify the denoising process by adding the gradient of $\mathcal{J}$ as guidance \citep{janner2022planning}:
\begin{equation}
p_{\theta}(\mathbf{z}^{k-1} \mid \mathbf{z}^{k}, \textit{\textbf{c}}) \approx \mathcal{N}(\mathbf{z}^{k-1}; \mu + \Sigma g, \Sigma)
\end{equation}
where $\mu = \mu_\theta$ as defined in (\ref{denoising_process}), and $g = \nabla \mathcal{J}(\mathcal{D}_\theta (\mathbf{z}, \textit{\textbf{x}}, \textit{\textbf{m}}))$.

To further enhance the robustness of the guidance mechanism, we adopt the reconstruction guidance (clean guidance) strategy \citep{rempe2023trace}, which perturbs the clean latent vector $\hat{\mathbf{z}}^0$ predicted by the network. This approach mitigates numerical instabilities, ensuring a more stable and controllable denoising process.

In detail, the objective function $\mathcal{J}(\boldsymbol{\tau})$ comprises three components: (i) $\mathcal{J}_{bv\_real}$, which ensures that non-adversarial vehicles behave realistically by preventing collisions with each other and avoiding off-road deviations; (ii) $\mathcal{J}_{adv\_real}$, which maintains the plausibility of the adversarial vehicle’s behavior by preventing it from colliding with non-adversarial vehicles or leaving the roadway; and (iii) $\mathcal{J}_{adv}$, which is generated by the LLM mentioned in the previous section and formulated to control the adversarial vehicle’s behavior to induce a collision with the ego vehicle based on a user query. By integrating these objectives, the proposed framework effectively balances realism while enabling the controllable generation of adversarial driving scenarios.

The collision penalty between each vehicle is defined as:
\begin{equation}
\text{veh\_coll\_pens}_{ij}(t) =
\begin{cases} 
1 - \frac{d_{ij}(t)}{p_{ij}}, & \text{if } d_{ij}(t) \leq p_{ij} \\
0, & \text{otherwise}
\end{cases}
\end{equation}
where $d_{ij}(t) = \|\mathbf{x}_i(t) - \mathbf{x}_j(t)\|$ denotes the Euclidean distance between vehicles $i$ and $j$ at time $t$, and $p_{ij} = r_i + r_j + d_{\text{buffer}}$ represents the collision threshold based on the vehicle radii and a safety buffer.

Similarly, the map collision penalty for vehicles is given by:
\begin{equation}
\text{env\_coll\_pens}_{i}(t) =
\begin{cases} 
1 - \frac{d_{i}(t)}{p_{i}}, & \text{if } d_{i}(t) \leq p_{i} \\
0, & \text{otherwise}
\end{cases}
\end{equation}
where $d_{i}(t) = \|\mathbf{x}_i(t) - \mathbf{c}_i(t)\|$ is the distance from the vehicle center to the nearest non-drivable point at time $t$, and $p_{i}$ denotes the maximum allowable displacement before a collision.

In practice, $\mathcal{J}_{bv\_real}$ and $\mathcal{J}_{adv\_real}$ compute the cumulative loss for their respective vehicle types using the aforementioned penalty functions. The loss function is computed separately for non-adversarial and adversarial vehicles, with corresponding latent values updated independently. Specifically, the gradient update of $\mathcal{J}_{bv\_real}$ affects the latent representations of non-adversarial vehicles, while the gradient update of the sum of $\mathcal{J}_{adv\_real}$ and $\mathcal{J}_{adv}$ is applied to the latent variables corresponding to the adversarial vehicle, ensuring targeted control over its behavior.

\section{Experimental Results}
\label{sec:experiment}
This section outlines the evaluations of our proposed LD-Scene model. We first describe the dataset, evaluation metrics, and experimental setup. We then compare the performance of our LD-Scene against baseline models and provide a quantitative analysis of the results. Additionally, we perform two ablation studies: one to demonstrate the effectiveness of the guidance components and another to validate the effectiveness of the debugger module. Finally, we perform two controllability studies, examining both the controllable adversarial level and controllable adversarial behavior.

\subsection{Dataset}
We conduct our experiments on the nuScenes \citep{nuscenes2019} dataset, which comprises 1,000 driving scenes, each lasting 20 seconds and recorded at 2\,Hz. The dataset captures 5.5 hours of urban driving data collected from two cities, Boston and Singapore, covering diverse traffic conditions and complex interactions between road agents. We train our models using the training split and evaluate them on the validation split of nuScenes. Following the standard guidelines of the nuScenes prediction challenge, we use 2 seconds (4 steps) of past motion data to predict the future 6 seconds (12 steps) of trajectories.

\subsection{Evaluation Metrics}
We focus on realistic safety-critical scenarios and propose a set of metrics to evaluate the quality of the generated scenarios, mainly in three aspects: adversariality, Behavior Plausibility, and efficiency.
\begin{itemize}
    \item \textbf{Adversariality}: This aspect measures the effectiveness of the generated scenarios in simulating safety-critical situations for the ego vehicle. Adv-Ego Collision Rate quantifies the percentage of scenarios where the adversarial vehicle collides with the ego vehicle, with higher values indicating stronger adversarial effectiveness. Adv Acceleration (Adv Acc) measures the acceleration magnitude of the adversarial vehicle, where higher values represent more aggressive adversarial behavior.

    \item \textbf{Behavior Plausibility}: To ensure the generated scenarios align with real-world traffic behavior, we assess both offroad rates and collision rates. Adv Offroad Rate and Other Offroad Rate measure the frequency at which adversarial and non-adversarial agents drive offroad, where lower values indicate more plausible behavior. Collision rates, including Adv-Other Coll Rate, Other-Ego Coll Rate, and Other-Other Coll Rate, quantify the frequency of crashes among different vehicle groups, with lower values ensuring realistic interactions. 

    \item \textbf{Efficiency}: We use the closed-loop simulation time (Sim Time) as the metric, where a shorter simulation time indicates higher efficiency.
\end{itemize}

\subsection{Experimental Setup}
\textbf{Baseline.} We evaluate our proposed method against existing baseline approaches for generating adversarial driving scenarios. Specifically, we compare with: our re-implementation of AdvSim \citep{advsim}, which optimizes the acceleration of a predefined adversarial vehicle to induce a collision, with initial states generated by SimNet \citep{bergamini2021simnet}; and Strive \citep{strive}, for which we utilize its open-source implementation. Strive employs a learned traffic model and performs adversarial optimization in the latent space. We further compare with two diffusion-based baselines: DiffScene \citep{xu2023diffscene}, which incorporates a human-designed safety guidance function to guide the generation toward safety-critical scenarios; and Safe-Sim \citep{chang2024safe}, which enhances the guidance function by explicitly incorporating time-to-collision (TTC) constraints to generate more targeted adversarial scenarios.

\begin{table*}[!h]
\footnotesize
  \centering
  \renewcommand{\arraystretch}{1} 
  \setlength{\tabcolsep}{3.5pt}
  % \resizebox{\linewidth}{!}{
  \begin{tabular}{c|cc|ccccc|c}
    \toprule 
    \centering \multirow{2}{*}[-1.0ex]{Model} & \multicolumn{2}{c|}{\textbf{Adversariality}} & \multicolumn{5}{c|}{\textbf{Behavior Plausibility}} & \textbf{Efficiency} \\
    & \makecell{Adv-Ego \\ Coll (\%) $\uparrow$} 
    & \makecell{Adv \\ Acc (m/s$^2$) $\uparrow$}
    & \makecell{Adv \\ Offroad (\%) $\downarrow$}  
    & \makecell{Other \\ Offroad (\%) $\downarrow$}  
    & \makecell{Adv-Other \\ Coll (\%) $\downarrow$}   
    & \makecell{Other-Ego \\ Coll (\%) $\downarrow$}  
    & \makecell{Other-Other \\ Coll (\%) $\downarrow$} 
    & Sim Time (s) $\downarrow$ \\
    \midrule
    AdvSim  & 24.72 & 0.90 & 15.60 & \textbf{14.85} & \textbf{0.56} & \textbf{0.91} & 0.11 & 338.35 \\
    Strive  & 22.69 & 0.88 & 18.94 & 16.64 & 0.90 & 1.08 & \textbf{0.05} & 609.72\\
    DiffScene  & 15.06 & 0.98 & 19.71 & 19.65 & 8.03 & 2.60 & 1.67 & 199.01\\
    Safe-Sim  & 27.81 & 1.09 & 21.79 & 18.12 & 7.52 & 3.21 & 0.66 & \textbf{193.59} \\
    LD-Scene    & \textbf{40.75} & \textbf{1.36} & \textbf{12.52} & 17.95 & 4.93 & 2.17 & 0.66 & 229.40 \\
    \bottomrule
  \end{tabular}
  \caption{
  Overall performance comparison of baseline models on the nuScenes dataset. We compare our approach against AdvSim \citep{advsim}, Strive \citep{strive}, DiffScene \citep{xu2023diffscene}, and Safe-Sim \citep{chang2024safe} for closed-loop safety-critical traffic simulation with a rule-based planner.
  }
  \label{tab:overall_performance}
\end{table*}

\textbf{Implementation Details.} Our LD-Scene is implemented using the PyTorch framework and trained on four GeForce RTX 4090 GPUs for six hours. The diffusion model is trained for 200 epochs using the Adam optimizer with a learning rate of $5 \times 10^{-4}$. The number of diffusion steps is set to 20. During the inference stage, we generate multiple candidate future trajectories for each non-ego agent in a given scene, with the number of test samples set to 10. The final trajectory is selected as the one that minimizes the guidance loss and simultaneously satisfies the physical feasibility constraints \citep{peng2025safety}. Additionally, both the code generator and debugger used in generating guidance utilize the GPT-4o model.

To ensure a fair evaluation of the controllability of the generated safety-critical scenarios, we adopt a standardized strategy for selecting the adversarial vehicle for each model. Specifically, we assign the adversarial vehicle as the one closest to the ego vehicle in the initial state of the scenario and ensure that it satisfies the required feasibility conditions. This selection strategy follows the approach used in Strive \citep{strive}; however, unlike Strive, the adversarial vehicle remains fixed throughout the scenario and does not change dynamically. 
Moreover, all experiments are conducted under a closed-loop simulation setup, where the ego vehicle is controlled by a rule-based lane-graph planner \citep{montemerlo2008junior}, ensuring consistent and reactive behavior throughout the scenario.

\begin{figure*}
\centerline{\includegraphics[width=0.98\linewidth, trim= 0 400 00 10, clip]{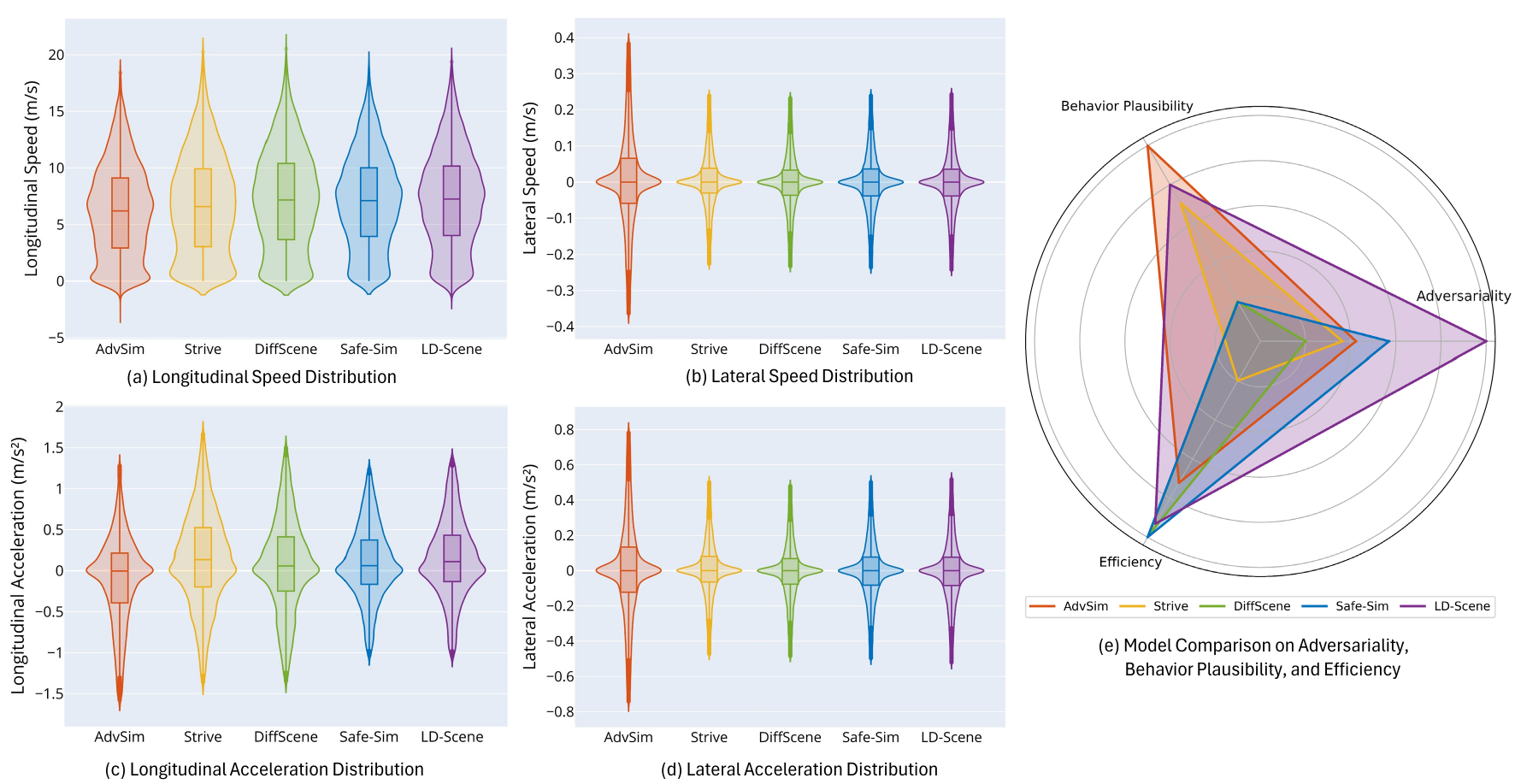}}
\caption{
Visualizations of overall performance among scenario generation models. (a)–(d): Distributions of longitudinal/lateral speed and acceleration. (e): Radar chart comparing adversariality, behavior plausibility, and generation efficiency. LD-Scene achieves the best overall balance.
}
\label{fig:model_compare}
\end{figure*}

\subsection{Overall Performance}
The quantitative performance of our approach and the baselines is presented in Table~\ref{tab:overall_performance}. Compared to the baselines, the proposed LD-Scene demonstrates significant advantages in adversarial metrics while maintaining realistic and plausible agent behaviors. Specifically, our LD-Scene achieves an Adv-Ego Coll (\%) of 40.75\%, which is substantially higher than the baseline models. This highlights the superior capability of our model in controllably generating safety-critical driving scenarios that involve adversarial vehicles colliding with the ego vehicle. In terms of realism, the LD-Scene shows a lower Adv Offroad (\%) of 12.52\% compared to AdvSim (15.60\%), Strive (18.94\%), DiffScene (19.71\%), and Safe-Sim (21.79\%), indicating our model can better generate adversarial scenarios that are realistic and feasible within the road constraints. The LD-Scene also maintains an overall good performance across other metrics in realism. For instance, the Other Offroad (\%) is 17.95\%, which is comparable to the baseline models. Additionally, the collision rates involving other vehicles (Adv-Other Coll, Other-Ego Coll, and Other-Other Coll) are all within reasonable ranges, further confirming the realistic nature of the generated scenarios. Furthermore, our diffusion-based framework improves generation efficiency, with LD-Scene achieving an average inference time of 229.40 seconds, significantly faster than test-time optimization methods.

Moreover, Fig.~\ref{fig:model_compare}(a)-(d) illustrate the driving behavior distributions of non-adversarial agents. Our LD-Scene exhibits more concentrated and narrower distributions, indicating smoother and more stable driving behavior compared to the baseline models. In particular, the narrower lateral speed and acceleration distributions show that LD-Scene avoids unreasonable sharp turns, while the compact longitudinal distributions reflect stable speed regulation. Fig.~\ref{fig:model_compare}(e) presents a radar chart summarizing each model’s performance in adversariality, behavior plausibility, and efficiency. Here, LD-Scene achieves the highest adversariality score, demonstrating its superior ability to induce safety-critical events, and also records strong behavior plausibility and efficiency values. This balanced performance confirms that LD-Scene achieves the strongest adversarial effectiveness while maintaining high levels of realism and generation efficiency, thus demonstrating its superiority across all evaluation metrics.

\begin{figure*}[!h]
\centerline{\includegraphics[width=0.95\linewidth]{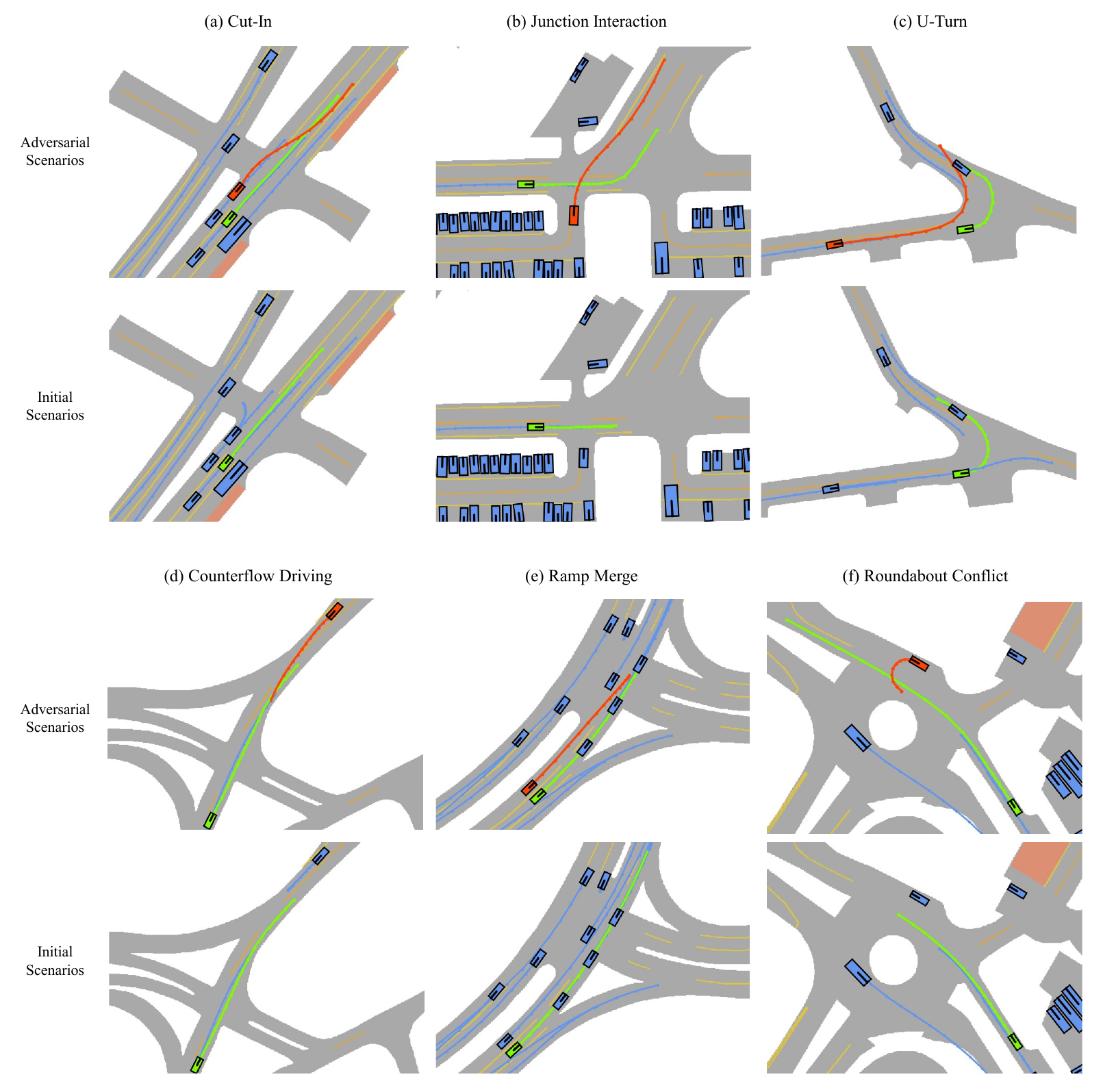}}
\caption{Adversarial safety-critical scenarios generated by LD-Scene. Examples include cut-in maneuvers, junction interactions, U-turns, counterflow driving, ramp merges, and roundabout conflicts, demonstrating our model’s capability to generate diverse and realistic safety-critical scenarios.}
\label{fig:adv_example}
\end{figure*}

Fig.~\ref{fig:adv_example} shows several example adversarial safety-critical scenarios generated by our proposed LD-Scene. In each subfigure, the red vehicle represents the adversarial vehicle, while the green vehicle is the ego vehicle controlled by a rule-based planner. Under varying scene contexts, the adversarial vehicle exhibits a range of aggressive behaviors, ultimately resulting in collisions with the ego vehicle in closed-loop simulations. This highlights the effectiveness of LD-Scene in generating safety-critical scenarios that expose the limitations and vulnerabilities of autonomous driving systems. 
For example, Fig.~\ref{fig:adv_example}(b) and Fig.~\ref{fig:adv_example}(e) illustrate typical scenarios where the ego vehicle encounters a critical decision between yielding and passing. Such scenarios test the planner’s ability to handle interactions, which remain a key concern in real-world autonomous driving. Furthermore, scenarios like Fig.~\ref{fig:adv_example}(d) and Fig.~\ref{fig:adv_example}(f) involve violations of traffic rules. These scenarios are particularly difficult to anticipate and handle, but their inclusion is crucial for improving the robustness of autonomous driving systems.

Overall, these results demonstrate the capability of our framework to generate realistic safety-critical driving scenarios, which can serve as valuable stress tests for autonomous vehicle systems.

\subsection{Ablation Study}
\subsubsection{Effectiveness of the Guidance Components}
Table~\ref{tab:guidance_ablation} demonstrates the contribution of each guidance component to the overall generation performance. Specifically, with the integration of all three guidance losses (Other-real Guidance, Adv-real Guidance, and Adv Guidance), the model shows superior performance compared to a baseline diffusion model without any guidance. The Adv Guidance component is particularly critical for increasing the collision rate between an adversarial vehicle and the ego vehicle, achieving an Adv-Ego Coll (\%) of 39.68\%, far exceeding the baseline (1.26\%). Additionally, the Adv-real Guidance lowers the probability of adversarial vehicles going off-road, reducing the Adv Offroad (\%) from 13.15\% to 10.68\%, thereby enhancing the realism and feasibility of adversarial trajectories. Meanwhile, the Other-real Guidance improves the feasibility of non-adversarial vehicle trajectories by decreasing the Other Offroad (\%) to 17.95\% and maintaining lower collision rates involving other vehicles, such as Adv-Other Coll (\%) at 4.93\% and Other-Other Coll (\%) at 0.66\%. Consequently, the proposed LD-Scene achieves an optimal Adv-Ego Coll (\%) of 40.75\% while keeping the Adv Offroad (\%) at a reasonable level of 12.52\%, confirming its effectiveness in generating both highly adversarial and realistic safety-critical driving scenarios.

\begin{table*}
\footnotesize
  \centering
  \renewcommand{\arraystretch}{1} % Adjust row spacing
  \setlength{\tabcolsep}{5.2pt} % Adjust column spacing
  % \resizebox{\linewidth}{!}{
  % \begin{tabular}{ccc|c|c|c|c|c}
  \begin{tabular}{>{\centering\arraybackslash}p{1.28cm}
                  >{\centering\arraybackslash}p{1.25cm}
                  >{\centering\arraybackslash}p{1.25cm}|
                  >{\centering\arraybackslash}p{1.8cm}|
                  >{\centering\arraybackslash}p{1.8cm}|
                  >{\centering\arraybackslash}p{1.8cm}|
                  >{\centering\arraybackslash}p{1.8cm}|
                  >{\centering\arraybackslash}p{1.8cm}}
    \toprule
    \makecell{Other-real \\ Guidance} & \makecell{Adv-real \\ Guidance} & \makecell{Adv \\ Guidance} & \makecell{Adv-Ego \\ Coll (\%) $\uparrow$} & \makecell{Adv \\ Offroad (\%) $\downarrow$} & \makecell{Other \\ Offroad (\%) $\downarrow$} & \makecell{Adv-Other \\ Coll (\%) $\downarrow$} & \makecell{Other-Other \\ Coll (\%) $\downarrow$} \\
    \midrule
     $\times$ &  $\times$ &  $\times$ &  1.26  &  11.44  &  20.24  &  6.12  &  2.86  \\
     $\times$ &  $\times$ &  $\checkmark$ &  39.68 &  13.15  &  19.56  &  \textbf{4.70}  &  1.23  \\
     $\times$ &  $\checkmark$ &  $\checkmark$ &  39.18 &  \textbf{10.68}  &  18.66  &  5.30  &  0.96 \\
     $\checkmark$ &  $\checkmark$ &  $\checkmark$ &  \textbf{40.75}  &  12.52  &  \textbf{17.95}  &  4.93  &  \textbf{0.66}  \\
    \bottomrule
  \end{tabular}
  \caption{Ablation study on guidance settings. This table presents the impact of different guidance settings on model performance. The study evaluates the effects of different guidance loss components on adversarial effectiveness and realism in safety-critical traffic scenarios.}
  % }
  \label{tab:guidance_ablation}
\end{table*}

\begin{figure*}[!h]
\centerline{\includegraphics[width=0.98\linewidth, trim= 15 200 0 0, clip]{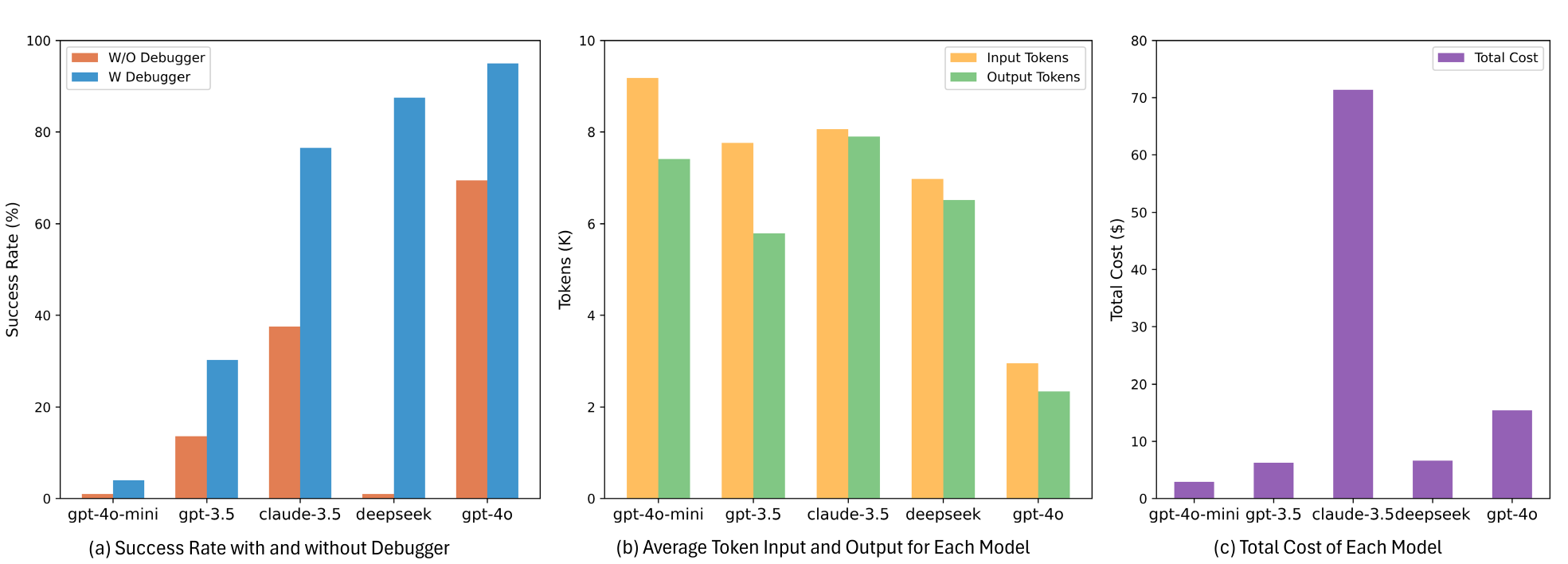}}
\caption{
Ablation study on the effectiveness of the debugger module. (a): Success rates of guidance code execution with and without the debugger. The debugger significantly improves success rates across all LLMs. (b): Average token input and output statistics per model. (c): Total cost of each model. 
}
\label{fig:debuger_exp}
\end{figure*}

\subsubsection{Effectiveness of the Debugger Module}
To validate the stability improvements afforded by the debugger module in our LD-Scene, we compare the execution success rates of the generated guidance functions for different LLM models under both with and without debugger settings. We evaluate a total of 500 user queries, which are automatically generated by GPT-4o. As shown in Fig.~\ref{fig:debuger_exp}(a), the debugger significantly improves success rates for all evaluated LLMs. For example, GPT-4o’s success rate increases from 69.4\% to 95.0\%, achieving highly stable generation. These results confirm that the debugger module can effectively detect and correct generation errors, ensuring reliable guidance function generation. In addition, we assess token consumption and total cost for different LLM models in Fig.~\ref{fig:debuger_exp}(b) and Fig.~\ref{fig:debuger_exp}(c), respectively. These results confirm that the debugger module offers substantial reliability improvements at a reasonable computational and financial expense. These insights further provide practical guidance for researchers in selecting suitable LLM models for similar tasks.

\subsection{Controllability Study}

\subsubsection{Controllable Adversarial Level}
We further investigate the controllability of our LD-Scene in terms of generating safety-critical driving scenarios with varying levels of adversarial intensity. Three adversarial levels, \textit{weak}, \textit{medium}, and \textit{strong}, are introduced for the study, and the three user query examples used in this experiment are as follows:
\begin{itemize}
    \item \textit{Weak}: Generate a guidance function class where the adversarial vehicle collides with the ego vehicle at a low speed during the interaction between the two vehicles.
    \item \textit{Medium}: Generate a guidance function class where the adversarial vehicle attempts to collide with the ego vehicle in a realistic and physically feasible manner.
    \item \textit{Strong}: Generate a guidance function class that encourages the adversarial vehicle to overtake the ego vehicle and collide with the ego vehicle at high speed.
\end{itemize}

\begin{figure*}
\centerline{\includegraphics[width=0.98\linewidth, trim= 0 195 0 0, clip]{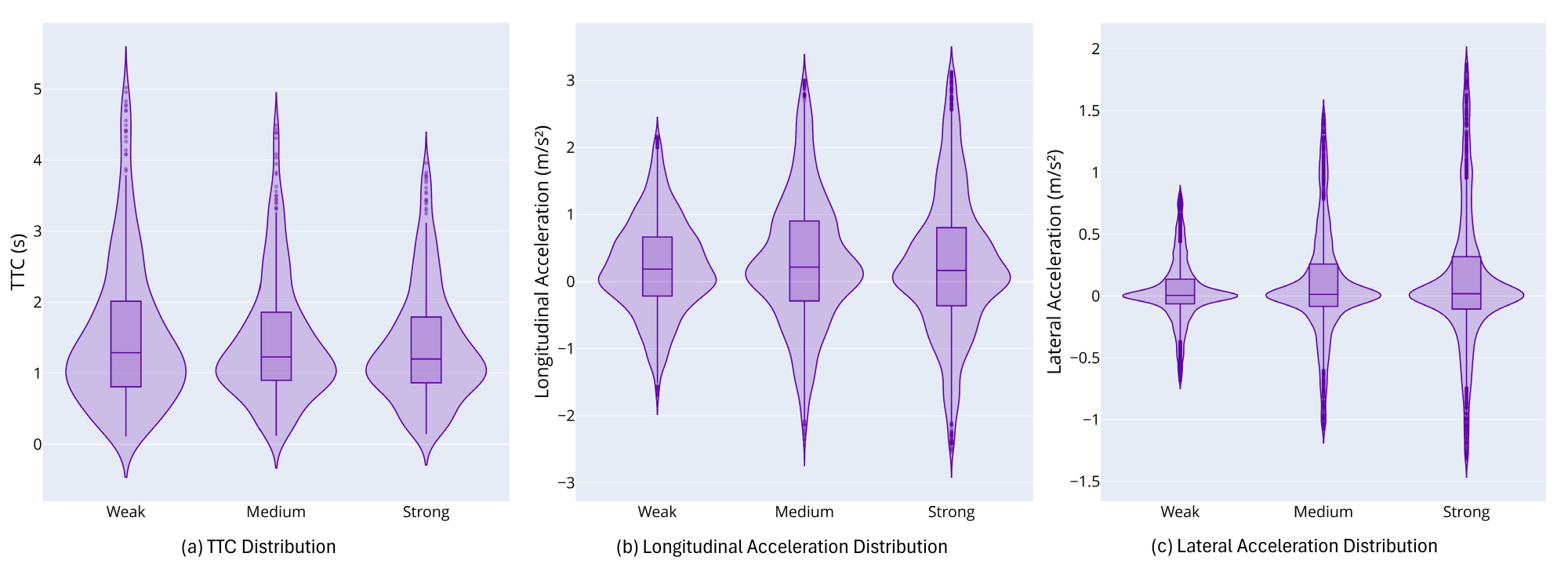}}
\caption{
Visualization of results under different adversarial levels.
}
\label{fig:level_compare}
\end{figure*}

The performance comparison under different adversarial levels is presented in Table~\ref{tab:adversarial_levels}, which demonstrates that our model can infer the desired adversarial intensity from the user query and generate corresponding safety-critical driving scenarios. At the weak adversarial level, the metrics such as TTC, Adv Acc\_lon, and Adv Acc\_lat indicate less aggressive driving behaviors. In contrast, both the Medium and Strong levels exhibit progressively more aggressive maneuvers, as evidenced by shorter TTC and increased acceleration values, thereby validating the controllability of adversarial behavior intensity. 
As illustrated in Fig.~\ref{fig:level_compare}(a), the TTC distribution shows progressively narrower peaks and lower medians from weak to strong levels. Meanwhile, Fig.~\ref{fig:level_compare}(b) and Fig.~\ref{fig:level_compare}(c) present the longitudinal and lateral acceleration distributions, respectively, which exhibit increasingly broader spreads toward higher acceleration values. These visualizations more clearly demonstrate the model’s ability to modulate scenario aggressiveness across multiple safety-critical dimensions. 
An additional observation is that the Adv-Ego Coll (\%) at the Strong level(39.33\%) is comparable to that at the medium level (40.75\%), showing that the strong level does not necessarily lead to a higher collision rate between the adversarial vehicle and ego vehicle across the NuScenes dataset. This could be because high-speed overtaking maneuvers may not always result in increased collision opportunities, especially in low-speed interaction scenarios where the timing for generating adversarial collisions can be easily missed. 

In summary, our LD-Scene provides effective control over the adversarial level, allowing users to generate safety-critical driving scenarios tailored to their specific needs. From weak to strong adversarial levels, the model serves as a practical tool for testing the performance and safety of autonomous driving systems under diverse risk conditions.

\begin{table*}
\centering
\renewcommand{\arraystretch}{1} % Adjust row spacing
\footnotesize
\setlength{\tabcolsep}{14pt} % Adjust column spacing
% \resizebox{\linewidth}{!}{
% \begin{tabular}{c|c|c|c|c|c}
\begin{tabular}{>{\centering\arraybackslash}m{2.1cm}|
                >{\centering\arraybackslash}m{1.3cm}|
                >{\centering\arraybackslash}m{1.3cm}|
                >{\centering\arraybackslash}m{1.65cm}|
                >{\centering\arraybackslash}m{1.65cm}|
                >{\centering\arraybackslash}m{1.65cm}}
\toprule
Adversarial Level & Adv-Ego Coll (\%) & TTC (s) & Adv Acc\_lon (m/s$^2$) & Adv Acc\_lat (m/s$^2$) &  Adv Offroad (\%) \\
\midrule
Weak  & 30.63 & 2.06 &  1.02 & 0.69 & 10.81 \\
Medium & 40.75 & 1.98 & 1.09 & 0.83 & 12.52 \\
Strong & 39.33 & 1.91 & 1.20 & 0.92 & 14.53 \\
\bottomrule
\end{tabular}
\caption{
Performance comparison under different adversarial levels.
}
% }
\label{tab:adversarial_levels}
\end{table*}

\begin{figure*}
\centerline{\includegraphics[width=\linewidth, trim= 0 10 0 -10, clip]{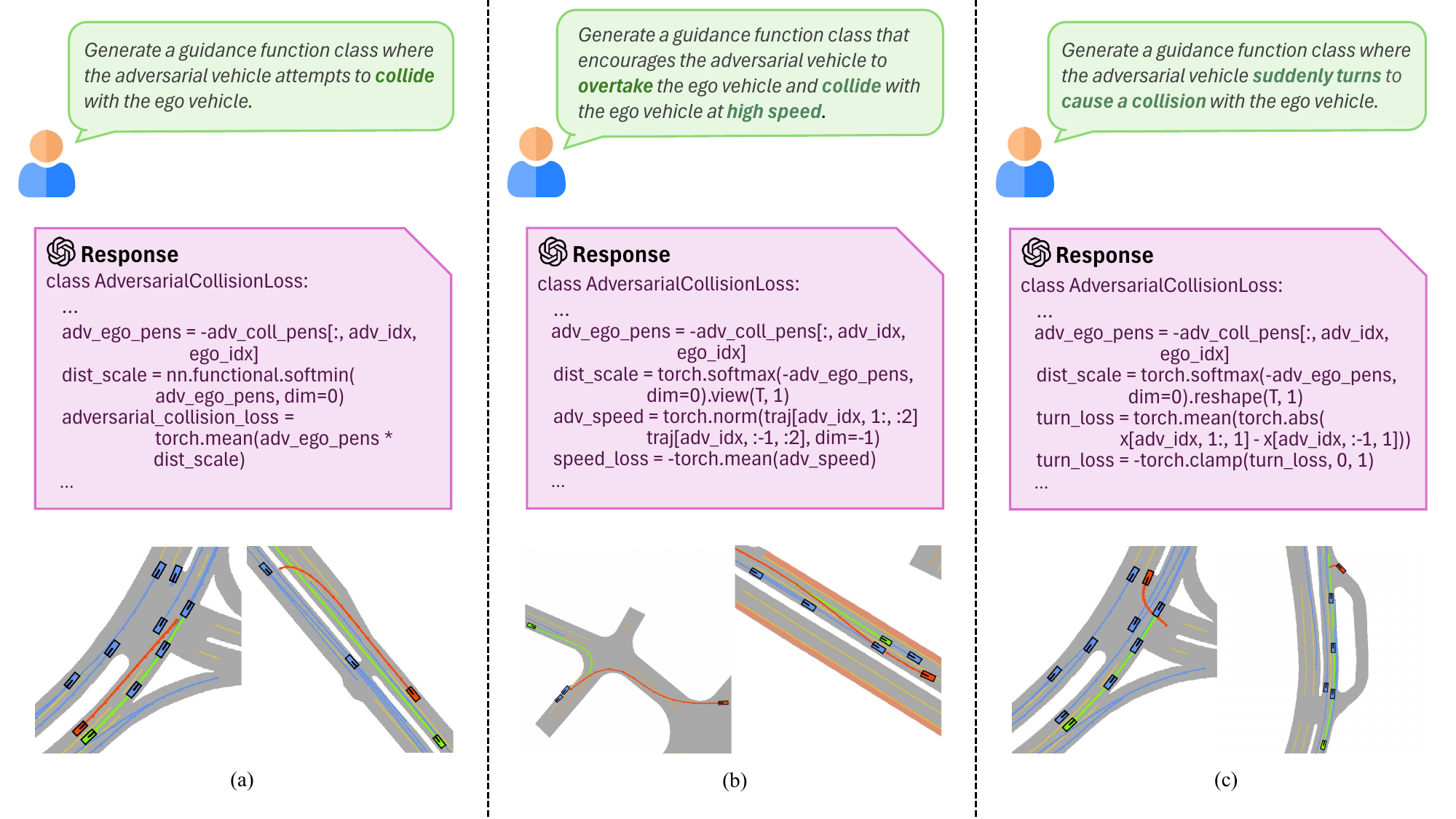}}
\caption{Case studies on adversarial safety-critical scenario generation based on different user queries, including normal collisions, high-speed overtaking collisions, and sharp-turn collisions. The results demonstrate the capability of our LD-Scene for controllable adversarial behavior, enabling the generation of diverse safety-critical scenarios.}
\label{fig:adv_control}
\end{figure*}

\subsubsection{Controllable Adversarial Behavior}
We conduct three case studies on different user queries for generating adversarial safety-critical scenarios, demonstrating our LD-Scene’s capability to synthesize diverse and controllable adversarial behaviors, as illustrated in Fig.~\ref{fig:adv_control}. Each case encompasses two example scenarios presented below the dialog. In case (a), the user query specifies a normal collision, leading LD-Scene to generate a loss function that minimizes the relative distance between the ego and adversarial vehicles. In case (b), the user query further includes a high-speed requirement, prompting LD-Scene to introduce an additional loss function to encourage high velocity. The associated scenarios illustrate the adversarial vehicle accelerating to overtake the ego vehicle before colliding, effectively simulating reckless high-speed maneuvers. Regarding case (c), the user query requests a sharp turn leading to a collision, and LD-Scene responds by generating a corresponding turn-related loss function. In both example scenarios, the adversarial vehicle abruptly swerves off its path, ultimately resulting in a collision. These results showcase that our LD-Scene can dynamically generate safety-critical scenarios based on different user queries, which provides a flexible and controllable framework for evaluating autonomous vehicle performance under diverse adversarial situations. 

\begin{figure*}
\centerline{\includegraphics[width=0.98\linewidth, trim= 10 280 10 10, clip]{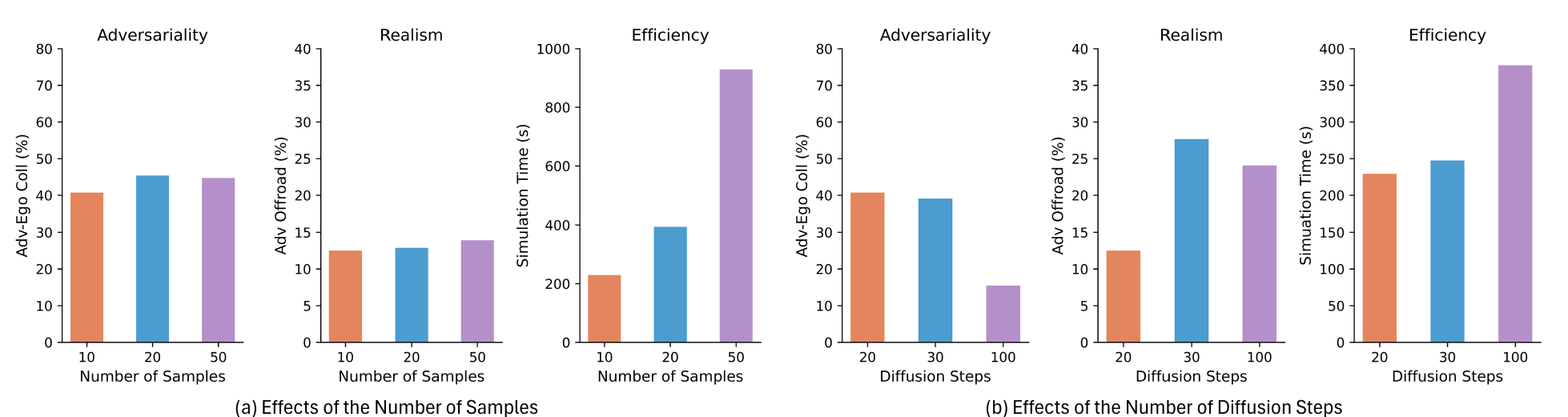}}
\caption{Quantitative results of key parameters on adversarial scenario generation, evaluated in terms of adversariality (Adv-Ego collision rate), realism (Adv Offroad collision rate), and efficiency (time consumption).}
\label{fig:param}
\end{figure*}

\subsection{Analysis of the Impact of Key Parameters}
We analyze the impact of two key parameters, namely the number of samples and the number of diffusion steps, on safety-critical scenario generation. The evaluation is based on three criteria: adversariality (Adv-Ego collision rate), realism (Adv Offroad collision rate), and efficiency (time consumption).

\textbf{Effects of the Number of Samples.}
The number of samples primarily controls the statistical stability and diversity of the generated adversarial scenarios. With an increased number of samples, as illustrated in Fig.~\ref{fig:param}(a), the overall performance exhibits slight improvement. However, this also leads to a significant rise in time consumption, resulting in reduced efficiency. Therefore, there is a clear trade-off between performance and computational cost. In this work, we choose the sample number as 10, which provides a good balance between generation quality and efficiency.

\textbf{Effects of the Number of Diffusion Steps.}
The number of diffusion steps controls the extent of denoising during the reverse diffusion process. A larger number of steps indicates more thorough denoising. However, as illustrated in Fig.~\ref{fig:param}(b), increasing the diffusion steps leads to a degradation in both adversariality and realism, while also incurring higher time consumption. This suggests that more thorough denoising does not necessarily yield better generation quality. In fact, excessive diffusion steps may cause error accumulation, ultimately harming performance. Based on this observation, we adopt 20 diffusion steps in our implementation, which achieves satisfactory results with acceptable computational cost.

\section{Conclusion}
\label{sec:conclusion}
In this paper, we present LD-Scene, an LLM-guided diffusion framework for the controllable generation of adversarial safety-critical driving scenarios. By integrating LDMs with LLM-enhanced guidance, our approach enables flexible, user-friendly scenario generation while ensuring both realism and adversarial effectiveness. Our method leverages CoT reasoning for structured guidance loss generation and incorporates an automated debugging module to enhance the reliability and stability of the generated guidance. Extensive experiments conducted on the nuScenes dataset demonstrate that our LD-Scene outperforms existing adversarial scenario generation baselines in terms of adversarial effectiveness, realism, and efficiency. Moreover, our framework allows natural language-based customization, making it accessible to users without extensive domain expertise. The controllability studies further demonstrate that our LD-Scene effectively modulates both the adversarial level and some specific adversarial behaviors, facilitating a more rigorous evaluation of AV performance under diverse safety-critical scenarios.

% \section*{Acknowledgement}

\bibliographystyle{cas-model2-names}

\bibliography{reference}

\end{document}